\documentclass{article}

\PassOptionsToPackage{numbers, compress}{natbib}
\usepackage[preprint]{neurips_2024}

\usepackage[utf8]{inputenc} % allow utf-8 input
\usepackage[T1]{fontenc}    % use 8-bit T1 fonts
\usepackage{hyperref}       % hyperlinks
\usepackage{url}            % simple URL typesetting
\usepackage{booktabs}       % professional-quality tables
\usepackage{amsfonts}       % blackboard math symbols
\usepackage{nicefrac}       % compact symbols for 1/2, etc.
\usepackage{microtype}      % microtypography
\usepackage{xcolor}         % colors

\usepackage{pifont}
\usepackage{xcolor}
\usepackage{threeparttable}
\usepackage{amssymb}
\usepackage{enumitem}
\usepackage{algorithm}
\usepackage{algpseudocode}
\usepackage{graphicx}
\usepackage{amsmath}
\usepackage{dsfont}
\usepackage{multirow} 
\usepackage{array}
\usepackage{wrapfig}

\title{Self-Calibrating 4D Novel View Synthesis from Monocular Videos Using Gaussian Splatting }

\author{%
  Fang Li,  Hao Zhang,  Narendra Ahuja \\
  University of Illinois at Urbana-Champaign\\
  \texttt{\{fangli3, haoz19, n-ahuja\}@illinois.edu} \\
}

\begin{document}

\maketitle

\begin{abstract}

Gaussian Splatting (GS) has significantly elevated scene reconstruction efficiency and novel view synthesis (NVS) accuracy compared to Neural Radiance Fields (NeRF), particularly for dynamic scenes. However, current 4D NVS methods, whether based on GS or NeRF, primarily rely on camera parameters provided by COLMAP and even utilize sparse point clouds generated by COLMAP for initialization, which lack accuracy as well are time-consuming. This sometimes results in poor dynamic scene representation, especially in scenes with large object movements, or extreme camera conditions e.g. small translations combined with large rotations. Some studies simultaneously optimize the estimation of camera parameters and scenes, supervised by additional information like depth, optical flow, etc. obtained from off-the-shelf models. Using this unverified information as ground truth can reduce robustness and accuracy, which does
%, and effectiveness but also increases the redundancy of the model structure. Such situations
frequently occur for long monocular videos (with e.g. $>$ hundreds of frames). We propose a novel approach that learns a high-fidelity 4D GS scene representation with self-calibration of camera parameters. It includes the extraction of 2D point features that robustly represent 3D structure, and their use for subsequent joint optimization of camera parameters and 3D structure towards
%. Then, we implement the canonical field and deformation field for the 
overall 4D scene optimization. We demonstrate the accuracy and time efficiency of our method through extensive quantitative and qualitative experimental results on several standard benchmarks. The results show significant improvements over state-of-the-art methods for 4D novel view synthesis. The source code will be released soon at \href{https://github.com/fangli333/SC-4DGS}{https://github.com/fangli333/SC-4DGS}.
\end{abstract}

\section{Introduction}
% \textcolor{red}{list, all the scope of our include all these things, assumptions, limitations, to the best of our knowledge}
This paper is about joint optimization of camera parameters and high-fidelity dynamic scene representation for photorealistic 4D novel view synthesis (NVS). Neural Radiance Field (NeRF)~\cite{nerf} approaches have significantly advanced the performance of Novel View Synthesis (NVS). However, NeRF-based methods~\cite{nerf, open-nerf, lerf, nerfies, hypernerf}, with their reliance on ray-casting and point-sampling algorithms, still suffer from long preprocessing and training times, which highly restrict their applications in areas like AR/VR and 3D content generation. Recently, 3D-GS~\cite{3dgs} has introduced explicit 3D representation and Differential-Gaussian-Rasterization in place of NeRF's implicit neural representation and neural renderer. Such improvements~\cite{3dgs,deformable-3dgs, 4dgs} have significantly reduced the training time while maintaining high-granularity rendering of novel views. It is important to note that the effectiveness of current NVS methods highly relies on the accuracy of camera parameters, which are obtained either from COLMAP\cite{colmap} or using self-calibration. The quality of these estimates affects the results obtained by various NVS methods.
\begin{table}[htbp]
  \caption{\textbf{Some Common Limitations and Indications of Whether Different Approaches Overcome them (\ding{51}) or Not (\ding{55}).} The total time shown is for camera \& scene optimizations on the \texttt{bell} data in NeRF-DS, using COLMAP docker~\cite{mini-colmap} indicated by \texttt{Col}.}
  \label{methodcomparisons}
  \centering
  \begin{threeparttable}
    \scalebox{0.69}{ % Adjust the scale factor as needed
      \begin{tabular}{ccccccccc}
        \toprule
        Method  & \begin{tabular}[c]{@{}c@{}}Dynamic \\ Scene \\NVS\end{tabular} &   \begin{tabular}[c]{@{}c@{}} Optimize\\ w/o G.T.\\ Intrinsic\end{tabular} &   \begin{tabular}[c]{@{}c@{}}Extreme \\Geometry \\Scene\end{tabular}  & \begin{tabular}[c]{@{}c@{}}Long-video\\Camera \\ estimation\end{tabular} & \begin{tabular}[c]{@{}c@{}}Accurate \\ Extracted\\Feature \end{tabular} & \begin{tabular}[c]{@{}c@{}}$\leq1$ more \\ Models' \\ Supervision\end{tabular} & \begin{tabular}[c]{@{}c@{}}Optimize\\w/o 3D \\ Priors\end{tabular} & \begin{tabular}[c]{@{}c@{}}Total\\ Optimization \\Time\end{tabular}\\
        \midrule
        NeRFmm~\cite{nerf--}                    & \ding{55}  & \ding{51} & \ding{55} & -         & \ding{55} & \ding{51} & \ding{51} & -             \\
        
        CF-3DGS~\cite{cf-3dgs}                  & \ding{55}  & \ding{55} & -         & -         & -         & \ding{51} & \ding{51} & -             \\
        
        Nope-NeRF~\cite{nope-nerf}              & \ding{55}  & \ding{55} & -         & -         & -         & \ding{51} & \ding{51} & -             \\
        
        LocalNeRF~\cite{localnerf}              & \ding{55}  & \ding{55} & -         & \ding{51} & -         & \ding{51} & \ding{51} & -             \\
        
        InstantSplat~\cite{instantsplat}        & \ding{55}  & \ding{51} & -         & -         & -         & \ding{51} & \ding{55} & -             \\
        
        FlowMap~\cite{flowmap}                  & \ding{55}  & \ding{51} & -         & \ding{51} & \ding{55} & \ding{55} & \ding{51} & -             \\
        
        \texttt{Col} $+$ HyperNeRF~\cite{hypernerf}              & \ding{51}  & \ding{51} & \ding{55} & \ding{51}   & \ding{55} & \ding{51} & \ding{51} & $>$64h        \\
        
        \texttt{Col} $+$ Deform-3DGS~\cite{deformable-3dgs}  & \ding{51}  & \ding{51} & \ding{55} & \ding{51} & \ding{55} & \ding{51} & \ding{51} & 5.7h \\
        
        \texttt{Col} $+$ 4D-GS~\cite{4dgs}                       & \ding{51}  & \ding{51} & \ding{55} & \ding{51} & \ding{55} & \ding{51} & \ding{51} & -             \\
        
        RoDynRF~\cite{rodynrf}                  & \ding{51}  & \ding{51} & \ding{51} & \ding{55} & \ding{55} & \ding{55} & \ding{51} & 30h           \\
        
        \textbf{Ours(SC-4DGS)}                  & \ding{51}  & \ding{51} & \ding{51} & \ding{51} & \ding{51} & \ding{51} & \ding{51} & 5h            \\
        \bottomrule
      \end{tabular}
      
    }
    % \scalebox{0.7}{
    % \begin{tablenotes}\small
    %   \item[a] The total time of preprocessing and camera \& scene optimizations on the \texttt{bell} data of NeRF-DS~\cite{nerfds}.
    %   \item[b] Usage of COLMAP docker~\cite{mini-colmap} with fewer features than the default due to our device limitation.
    %   \item[c] \texttt{Col} represents COLMAP~\cite{colmap}.
    % \end{tablenotes}
    % }
  \end{threeparttable}
  \vspace{-7pt}
\end{table}

Our goal is this paper is an NVS method that performs well with respect to eight properties. These are motivated by the limitations we have noted in the current methods. Specifically, we wish to overcome eight limitations (\texttt{L1-L8}) listed below.
\texttt{(L1)}: inaccurate feature extraction
%by SIFT~\cite{sift};
\texttt{(L2)}: long training time;
\texttt{(L3)}: works only for static scenes, does not extend to dynamic scenes;
\texttt{(L4)}: does not work under extreme geometric conditions - having large object movements and camera rotations but small camera translations such as DAVIS\cite{davis} dataset;
\texttt{(L5)}: requires specification of camera intrinsics;
\texttt{(L6)}: requires specification of 3D prior model~\cite{dust3r};
\texttt{(L7)}: requires supervision from multiple models (to overcome their individual limitations);
and
\texttt{(L8)}: does not work well on long videos.

Examples of past work whose limitations motivate specific properties are as follows.
 NeRFmm~\cite{nerf--} requires the camera to forward-face the scenes and the rotation range of the camera be limited to $\pm 20^\circ$ (\texttt{L4}).
 Nope-NeRF~\cite{nope-nerf}, CF-3DGS\cite{cf-3dgs} and LocalNeRF\cite{localnerf} lack (\texttt{L5}) and also require monocular depth estimation from MiDaS~\cite{midas}.
 Nope-NeRF also needs about 30 hours of optimization for each scene (\texttt{L2}).
 InstantSplat~\cite{instantsplat} cannot work without the 3D prior model DUSt3R~\cite{dust3r} in (\texttt{L6}).
 FlowMap~\cite{flowmap} requires off-the-shelf models RAFT~\cite{raft}, MiDaS~\cite{midas}, and CoTracker~\cite{cotracker} for computing optical flow, monocular depth estimation, and point tracking, and suffers from the inaccuracies arising from these models; Sec \ref{ablation} shows failure cases. (\texttt{L7}) .

 All the aforementioned methods fail on dynamic scenes (\texttt{L3}). 4D-GS~\cite{4dgs}, Deformable 3DGS~\cite{deformable-3dgs} and HyperNeRF~\cite{hypernerf} methods address (\texttt{L3}) but depend on COLMAP~\cite{colmap} for camera parameter estimation in the absence of a good alternative for dynamic scenes. Even when they eliminate moving objects using motion masks, the performance of COLMAP and the 4D scene optimization models is still limited by (\texttt{L1}) and (\texttt{L2}), resulting in poor 4D NVS performance. Further, our experiments show that COLMAP~\cite{colmap} completely fails in the extreme geometry scenes with relatively large object movements and tiny camera translations but huge camera rotations such as seen in the DAVIS~\cite{davis} dataset (\texttt{L4}), consistent with the findings in RodynRF~\cite{rodynrf} (Sec \ref{cameraevaluation}). As a result of incorporating supervision from MiDaS~\cite{midas} and RAFT~\cite{raft}, RoDynRF~\cite{rodynrf} is affected by (\texttt{L2}), (\texttt{L7}) and (\texttt{L8}), requiring over 28 hours' training for one monocular video with 50$\sim$80 frames and failing while optimizing long videos (e.g. $>$ 800 frames).
 Tab \ref{methodcomparisons} and Fig \ref{render-nerfds} lists these limitations and how different methods compare with respect to them, where \ding{51} indicates that the limitation is overcome (desirable).

In comparison to this SOTA, in this paper we propose a new method \textbf{SC-4DGS} which can robustly learn accurate camera parameters and reconstruct high-fidelity dynamic scene representations, free of the limitations above. SC-4DGS starts with our proposed Structural Points Extraction (SPE) algorithm (Sec \ref{sec3.2}), which can extract highly accurate 2D-3D mappings of structural points based only on CoTracker~\cite{cotracker}, for accurate camera estimation.
%SPE includes an automatic structural point adaptation mechanism to ???????????????????perform well???????????????????? on monocular videos with large inter-frame movement and ???????????less repetition????????????????.
We then jointly optimize camera parameters and 3D structural points supervised by the extracted 2D structural points and 2D-3D correspondence from SPE (Sec \ref{sec3.3}). Finally, with the optimized camera parameters and 3D structural points, optimal scene representations are optimized within the canonical field (to learn mean positions $x$, mean quaternions $r$, mean scaling $s$, and opacity $\sigma$) and the deformation field defined (to learn $\Delta x$, $\Delta r$, $\Delta s$) defined in Sec \ref{sec3.4}. We evaluate our approach on three standard public datasets including NeRF-DS~\cite{nerfds}, DAVIS~\cite{davis}, and Nvidia~\cite{nvidia}, and present quantitative and qualitative comparisons with existing methods.

Our \textbf{main contributions} are as follows:
\begin{itemize}[itemsep=1mm, topsep=0mm, parsep=0mm, leftmargin=8mm]
  \item We introduce a new method \textbf{SC-4DGS} that possesses a number of desirable properties: it synthesizes high-fidelity novel views of dynamic scenes using Gaussian Splatting without requiring camera priors and limitations on video length while taking less time than SOTA methods. Indeed, it overcomes all the limitations (\texttt{L1-L8}), as can be seen in Tab \ref{methodcomparisons}
  \item Towards the aforementioned performance, our SC-4DGS learns robust and accurate camera parameters, an ability whose lack has adversely affected the performances of many SOTA methods.
  %without getting stuck into each limitation mentioned in Tab.\ref{methodcomparisons}. 
  \item We show that our method outperforms the current state-of-the-art methods in quantitative and qualitative terms on three standard benchmark datasets.
  
\end{itemize}

\section{Related Work}
\label{relatedworks}

\textbf{Novel View Synthesis w/ COLMAP.} To reconstruct views of objects and scenes, existing methods employ different representations, including mesh representations~\cite{banmo, limr}, planar representations~\cite{appp, titp}, point cloud representations~\cite{pointnerf, point2}, neural field implicit representations~\cite{nerf, localnerf, hypernerf}, and the recently introduced explicit Gaussian representations~\cite{deformable-3dgs, 4dgs, 3dgs}. Prior to 3D-GS~\cite{3dgs}, numerous NeRF-based enhancements~\cite{nerf} were made, including dynamic scene synthesis~\cite{dnerf, hypernerf, nerfies, nerfds}, sparse-view scene reconstruction~\cite{infonerf, regnerf, sinnerf, freenerf}, and high-fidelity mesh extraction~\cite{banmo, neus, volsdf}. However, NeRF-based methods share the limitation of long training time. To address this issue, the recently introduced 3D-GS~\cite{3dgs} offers explicit 3D-GS representations and Differential-Gaussian-Rasterization rendering, implemented in CUDA~\cite{cuda}. This technique leverages learnable explicit 3D Gaussian ellipsoids, incorporating attributes like position, rotation, opacity, scale, and color for scene representations and reduce the time costs compared to NeRF-based methods. Several works~\cite{deformable-3dgs, 4dgs} have proposed its applications in dynamic scenes. Nonetheless, the efficiency and performance of both NeRF-based and 3D-GS-based methods are significantly hampered by the preprocessing time, and accuracy of camera parameters which are obtained from COLMAP~\cite{colmap}.

\textbf{Novel View Synthesis w/o COLMAP.} Currently, COLMAP~\cite{colmap} is the most widely used method for camera parameter estimation, but its limitations have been a barrier. The inaccuracies of its SIFT~\cite{sift} feature extraction, along with its time-consuming matching and reconstruction steps, have hindered its usage. Some methods~\cite{nerf--, barf, nope-nerf, cf-3dgs} attempt to jointly optimize camera poses and static scene representations, but they require camera intrinsics to be provided. InstantSplat~\cite{instantsplat} and FlowMap~\cite{flowmap} have been introduced to address such limitations in static scenes. In dynamic environments, COLMAP~\cite{colmap} is widely used by state-of-the-art dynamic scene NVS methods~\cite{hypernerf, deformable-3dgs, 4dgs} even though it is theoretically designed for static scenes - it optimizes camera parameters and 3D parameters in successive time steps and therefore cannot handle the changing 3D parameters in dynamic scenes. To overcome these challenges, RoDynRF~\cite{rodynrf} leverages supervision from monocular depth estimation~\cite{midas} and optical flow estimation~\cite{raft} in addition to RGB images. Unfortunately, RoDynRF struggles with long monocular videos and requires over 28 hours of training even for short videos. Compared with RoDynRF and other existing dynamic scene NVS methods utilizing COLMAP, our proposed method learns more accurate and robust camera parameters in less time without requiring any camera priors and produces comparable results.

\section{Method}

\begin{figure}[ht]
    \centering
    \includegraphics[width=\linewidth]{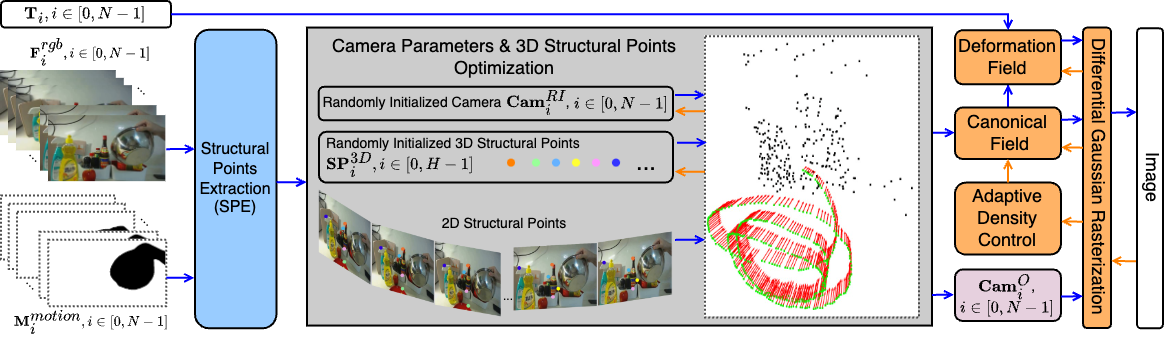}
    \vspace{-10pt}
    \small
    \caption{\textbf{Overview of SC-4DGS.} We take the \texttt{basin} data of NeRF-DS~\cite{nerfds} dataset as the example. First, the SPE algorithm attempts to extract 2D structural points $\mathbf{SP}^{2D}$ and 3D structural points $\mathbf{SP}^{3D}$ through $\mathbf{F}^{rgb}$ and $\mathbf{M}^{motion}$, and establish relationships among them. Then, the optimized cameras $\mathbf{Cam}^{O}$ are learned in the second joint optimization module, starting with randomly initialized camera $\mathbf{Cam}^{RI}$ and $\mathbf{SP}^{3D}$, supervised by the estimated $\mathbf{SP}^{2D}$ from SPE. Finally, given $\mathbf{T}$, the optimized $\mathbf{Cam}^{O}$ and $\mathbf{SP}^{3D}$, a Canonical Field and a Deformation Field (see text for details) are computed to optimize the mean representations and deformations of the scene, respectively, supervised by $\mathbf{F}^{rgb}$. In the middle of the figure, we show the learned camera positions \textcolor{green}{\textbullet} and orientations \textcolor{red}{$\rightarrow$}, and the optimized $\mathbf{SP}^{3D}$ \textcolor{black}{\textbullet}. \textcolor{blue}{$\rightarrow$} and \textcolor{orange}{$\leftarrow$} respectively represent operations flow and gradient flow.}
    \vspace{-7pt}
    \label{overview}
\end{figure}

We present an overview of our method in Fig \ref{overview}. Starting with a monocular video with $N$ frames, we input the RGB frames $\mathbf{F}_i^{rgb}$, motion masks $\mathbf{M}_{i}^{motion}$, and frame times $\mathbf{T}_{i}$ to our model, $i \in [0, N - 1]$. The steps in our method are presented in the subsections below. We first briefly review 3D Gaussian Splatting (3D-GS) in Sec \ref{sec3.1}. In Sec \ref{sec3.2}, we discuss our newly proposed Structural Point Extraction (SPE) algorithm, detailing how it can establish correspondences between 2D structural points in each frame and 3D structural points in world coordinates shared at successive frame times, and then extract them. Joint optimization of camera parameters and 3D structural points is presented in Sec \ref{sec3.3}. Finally, we discuss dynamic scene representation optimization in Sec \ref{sec3.4}.

\subsection{Preliminaries}
\label{sec3.1}

In significant contrast to the implicit scene representation in NeRF~\cite{nerf}, 3D-GS~\cite{3dgs} adopts a new way of representing the scene via explicit gaussian ellipsoids. Each 3D gaussian ellipsoid is parameterized by (a) center position $\mu \in \mathbb{R}^3$ in world coordinates; (b) quaternion (rotation) matrix $r \in \mathbb{R}^4$; (c)  opacity scalar $\alpha \in \mathbb{R}$; (d) scale factor $\sigma \in \mathbb{R}^{3}$, and (e) spherical harmonics (SH) coefficients $c \in \mathbb{R}^k$ ($k$ stands for the degree of freedom) representing the color by encoding the spatial distribution of light intensity across the surface of a sphere. The 3D gaussian ellipsoid $G$ is computed from its covariance matrix $\Sigma$ and its center $\mu$, and its 3D covariance matrix $\Sigma$ is calculated from the scaling factor $s$ and the quaternion matrix $r$ as in the following Eq \ref{eq1} and Eq \ref{eq2}:

\vspace{0pt}
\begin{equation}
\label{eq1}
\vspace{-0pt}
 G(x) = e^{-1/2(x - \mu)^T\Sigma^{-1}(x - \mu)} \\
\vspace{-0pt}
\end{equation}
\vspace{-15pt}

\begin{equation}
\label{eq2}
\vspace{-0pt}
\Sigma'= \mathbf{JW}\Sigma\mathbf{W}^{T}\mathbf{J}^{T}, \hspace{6pt} \Sigma = \mathbf{RS}\mathbf{S}^{T}\mathbf{R}^{T}
\vspace{-0pt}
\end{equation}

\vspace{0pt}

Given one 3D covariance matrix $\Sigma$ in world coordinates and the world-to-camera transformation matrix $\mathbf{W}$, the corresponding 2D covariance matrix $\Sigma'$ in camera coordinates is calculated as in Eq \ref{eq2}. $\mathbf{J}$ is the Jacobian of the affine approximation of the projection transformation. The color of each pixel $C(p)$ is rendered through the blending of the overlapped $K$ 2D gaussian ellipsoids on this pixel using volume rendering technique in Eq \ref{eq3}.

\vspace{-0pt}
\begin{equation}
\label{eq3}
\vspace{-0pt}
C(p) = \sum_{k \in K}c_{k}\alpha_{k}\Pi_{k}^{j - 1}(1 - \alpha_{k})
\vspace{-0pt}
\end{equation}
\vspace{-0pt}
Here, $c_{k}$ and $\alpha_{k}$ represent the color and density of this point calculated from $\Sigma$ multiplied by the opacity and SH color coefficients. Details can be seen in~\cite{3dgs, ewa}.

\subsection{Structural Points Extraction (SPE)}
\label{sec3.2}

\textbf{Terminology.} In this section, we define the variables used in the SPE algorithm to clarify the process. $\mathbf{F}_{i}^{rgb}$ and $\mathbf{M}_{i}^{motion}$ ($i \in [0, N - 1]$) represent the RGB frames and motion masks. $N$ is the number of frames in one monocular RGB video. $H$ is the total number of shared 3D structural points. $\mathbf{SP}^{3D} \in \mathbb{R}^{H \times 3}$, $\mathbf{SP}_{i}^{Pool} \in \mathbb{R}^{B \times 2}$, and $\mathbf{SP}_{i}^{2D} \in \mathbb{R}^{\tau \times 2}, i \in [0, N - 1]$ respectively represent the 3D structural points in the world coordinate, the potential 2D structural points, and the 2D structural points on each frame, with $\tau$ and $B$ are the number of required 2D structural points ($\tau = 100$ by default) and the number of the potential 2D structural points on each frame. $\mathbf{SP}_{i}^{Index}, i \in [0, N - 1]$ stores the mappings between 3D structural points and 2D structural points on each frame. $\mathbf{Cam}_{i}^{RI}$ and $\mathbf{Cam}_{i}^{O}$ are respectively the randomly initialized camera parameters and the optimized camera parameters. $\mathbf{F}_{i}^{gray}$ is the grayscale mapping of $F_{i}^{rgb}$. $\mathbf{Grad}_{i}^{magni}$ is the gradient magnitude obtained by combining the gradient norms of all three color channels $\mathbf{Grad}_{i}^{r}, \mathbf{Grad}_{i}^{g}, \mathbf{Grad}_{i}^{b}$ of $\mathbf{F}_{i}^{rgb}$. $\mathcal{W}$ is the window size of the Maximum Filter~\cite{opencv}. $\mathbf{E}_{i}^{gray} \in \{0, 1\}$ represents the edge detection information by Canny Edge Detector~\cite{canny} on $\mathbf{F}_{i}^{gray}$. $credit = \mathds{1}^{B}$ is to mark the remaining $\mathbf{SP}_{i}^{Pool}$ while processing frame by frame. $\mathbf{P}^{pos}$ and $\mathbf{P}^{index}$ are used to store the $\mathbf{SP}_{i}^{2D}$ and $\mathbf{SP}_{i}^{Index}$, and they are all initialized with $-1$ inside, representing '\texttt{TODO}'.

% \subsubsection{SPE} 
% \label{SPE}

\textbf{SPE algorithm} \ref{a1} begins by initializing $H=0$ and progressively increases $H$ by the number of newly introduced $\mathbf{SP}^{3D}$. While processing frames sequentially, given $\tau$, if $\mathbf{P}_{i}^{index}$ is still somehow 'empty' ($\mathbf{P}_{i}^{index}.any() \texttt{==} -1$), new $\mathbf{SP}_{i}^{2D}$ is selected from $\mathbf{SP}_{i}^{Pool}$ which stores the most 'representative' points on $\mathbf{F}_{i}^{rgb}$ into the following process. The reasons why we do not randomly select points on $\mathbf{F}_{i}^{rgb}$ and use CoTracker~\cite{cotracker} to track it are discussed in Sec \ref{ablation}. For the most 'representative' points, we first calculate $\mathbf{Grad}_{i}^{magni}$ of $\mathbf{F}_{i}^{rgb}$. $\mathbf{Grad}_{i}^{magni}$ can represent the frequency of each pixel through the gradient magnitude of the sum of the gradient norm in each color channel. However, although the local maxima of $\mathbf{Grad}_{i}^{magni}$ can tell us which pixels possess relatively high frequency, it still includes the pixels in the low-texture regions. Accurately tracking points on a low-texture surface is challenging for nearly all dense point-tracking models. To eliminate pixels in such low-texture regions, we use $\text{Canny Edge Detector}$ to obtain $\mathbf{E}_{i}^{gray}$ from the grayscale mapping $\mathbf{F}_{i}^{gray}$, according to the intensity across $\mathbf{F}_{i}^{rgb}$, and then do $\text{Maximum Filter}(\mathbf{E}_{i}^{gray} \cap \mathbf{Grad}_{i}^{magni}, \mathcal{W})$ to obtain the most 'representative' points across $\mathbf{F}_{i}^{rgb}$. Then, we use $\mathbf{M}_{i}^{motion}$ to mask out the potential 2D structural points on the deforming objects to acquire the final $\mathbf{SP}_{i}^{Pool}\in \mathbb{R}^{B \times 2}$, in that those are not appropriate for camera parameter estimation. $B$ can be different according to different $i$.

Given $\mathbf{SP}_{i}^{Pool}$, we use CoTracker~\cite{cotracker} to track the positions $\textbf{Pred}^{pos}$ and visibility $\textbf{Pred}^{vis}$ in the following frames. For $\textbf{SP}_{i}^{Pool}$ in $\textbf{F}_{i}^{rgb}$, if either the corresponding $\textbf{Pred}_{p}^{vis}\texttt{==}0$ or $\textbf{M}_{p}^{motion}[\textbf{Pred}_{p}^{pos}]\texttt{==}0$ ($p > i$), we mark these points as missing. When the number of remaining points first gets to be below  $num$, we randomly select $num$ points from the remaining points of last frame using $credit$, and store the corresponding points before the current frame into $\textbf{P}^{pos}$,  then assign them index in $\textbf{P}_{i}^{index}$. As the points continuously disappear in the following frames, we store only the existing points in each frame with their corresponding indexes. Our SPE algorithm performs such operations iteratively until all $\textbf{SP}_{i}^{2D}$ and $\textbf{SP}_{i}^{Index}$ are not 'empty'. The qualitative results of our SPE algorithm are shown in Fig \ref{ctspe} and Fig \ref{ctspe2}. More discussion on why we do not use the CoTracker output directly is in Sec \ref{ablation}.

\subsection{Camera Parameters \& 3D Structural Points Optimization}
\label{sec3.3}
Given $\mathbf{SP}^{2D}$, $\mathbf{SP}^{Index}$ and $\mathbf{SP}^{3D}$ from our SPE algorithm~\ref{a1} in Sec \ref{sec3.2}, we now discuss how to conduct joint optimization of camera parameters $\mathbf{Cam}$ and $\mathbf{SP}^{3D}$. We assume that the entire monocular video has a constant focal length $f$. We define the quaternion rotation matrix of $\mathbf{F}_{i}^{rgb}$ as $\mathbf{Quat}_{i} \in \mathbb{R}^{4}$, and parameterize orientation $\mathbf{R}_{i} \in \mathbb{R}^{3\times3}$, translation $\mathbf{T}_{i} \in \mathbb{R}^{3}$, world-to-camera transformation matrix $\mathbf{W2C}_{i} \in \mathbb{R}^{4\times4}$, and prospective projection matrix $\mathbf{PP} \in \mathbb{R}^{4\times4}$ following 3D-GS~\cite{3dgs}. With the mappings from 3D world coordinates to the pixel locations on the images in Eq \ref{eqcam}, we design the projection loss $\mathcal{L}_{proj} = \sum_{i = 0}^{N - 1}\text{MSE}(\mathbf{SP}_{i}^{2Dproj} - \mathbf{SP}_{i}^{2D})$\ between the 2D projected structural points $\mathbf{SP}_{i}^{2Dproj}$ and $\mathbf{SP}_{i}^{2D}$ from SPE \ref{a1}, the distance error loss $\mathcal{L}_{de} = \sum_{i = 0}^{N - 1}\text{MSE}(\text{Dist}(\mathbf{SP}_{i}^{2Dproj}) - \text{Dist}(\mathbf{SP}_{i}^{2D}))$ between the distances among $\mathbf{SP}_{i}^{2Dproj}$ and the ones from $\mathbf{SP}_{i}^{2D}$, and the depth regularization loss $\mathcal{L}_{dr} = \sum_{i = 0}^{N - 1}\text{ReLu}(-\mathbf{SP}_{i}^{Cam}[:, 3])$ for supervision in Eq \ref{eqeach}.
\vspace{5pt}
\begin{equation}
\label{eqcam}
\vspace{-0pt}
\mathbf{SP}_{i}^{Cam} = \text{Homo}(\mathbf{SP}^{3D}[\mathbf{SP}_{i}^{Index}])\mathbf{W2C}^{T}\mathbf{PP}^{T}, \hspace{4pt}
\mathbf{SP}_{i}^{2Dproj} = \mathbf{SP}_{i}^{Cam}[:, :2] / \mathbf{SP}_{i}^{Cam}[:, 3]
\end{equation}

\vspace{-18pt}

\begin{equation}
\label{eqeach}
\vspace{-0pt}
% \small
\begin{split}
    \mathcal{L}_{Cali} = \mathcal{L}_{proj} + \mathcal{L}_{de} + \mathcal{L}_{dr} \\
\end{split}
% \vspace{-0pt}
\end{equation}

Here, $\mathbf{SP}_{i}^{Cam} \in \mathbb{R}^{\tau \times 4}$, $\text{Dist}(\mathcal{X})$ represents the distance between each points $\mathcal{X}$. Homo converts the 3D points in the world coordinates into homogeneous coordinates by concatenating $\mathds{1}$. In total, we have $7N + 1 + H$ parameters required to be optimized in this step. Such supervision is shown to be effective for camera parameter estimation in Fig \ref{overview} and the discussions in Sec \ref{cameraevaluation}.

\subsection{Dynamic Scene Representations Optimization}
\label{sec3.4}
Unlike the existing 3D-GS-based methods~\cite{3dgs,4dgs,deformable-3dgs} which set up the initial point clouds either as the sparse point clouds from COLMAP~\cite{colmap} or as a cube full of random dense points, our SC-4DGS takes the optimized 3D structural points $\mathbf{SP}_{i}^{3DO}$ from Sec \ref{sec3.3} into the following dynamic scene representation optimization with Adaptive Density Control~\cite{deformable-3dgs}. We implement a Canonical Field $\mathcal{G_{C}}$~\cite{deformable-3dgs} and a Deformation Field $\mathcal{G_{D}}$~\cite{4dgs, deformable-3dgs} to learn the canonical scene representations $x, r, s, \sigma$ and deformation scene representations $\Delta x, \Delta r, \Delta s$ respectively as Eq \ref{eqoverall}.
\vspace{0pt}
\begin{equation}
\label{eqoverall}
\vspace{-0pt}
\begin{split}
  \Delta x_{i}, \Delta r_{i}, \Delta s_{i} = \mathcal{G}_{D}( \mathcal{P}(\mathcal{X}_{\mathcal{G}_{C}}), \mathcal{P}(\mathbf{T}_{i})) \\
\mathbf{I}_{i}^{Render} = \mathcal{R}(x + \Delta x_{i}, r + \Delta r_{i}, s + \Delta s_{i}, \sigma) \\  
\end{split}
\vspace{-0pt}
\end{equation}
\vspace{0pt}

In Eq \ref{eqoverall}, $\mathcal{X}_{\mathcal{G}_{C}}$ represents the learnt gaussian centers from $\mathcal{G}_{C}$, $\mathcal{X}_{\mathcal{G}_{C}} \leftarrow \mathcal{G}_{C}(\mathbf{SP}^{3D})$; $\mathcal{P}$ stands for the positional encoding on $\mathcal{X}_{G_{C}}$ and time $\mathbf{T}_{i}$, following~\cite{nerf, deformable-3dgs}. Besides, for simplicity, we use $\mathcal{R}$ standing for the Differential Gaussian Rasterization~\cite{3dgs} to render the image with main parameters, and using RGB loss~\cite{3dgs} $\mathcal{L}_{RGB} = (1 - \lambda)\mathcal{L}_{1} + \lambda\mathcal{L}_{D-SSIM}, \lambda = 0.2$ for supervision.

\vspace{-5pt}
\section{Experiments}
\vspace{-3pt}
COLMAP~\cite{colmap} and the recently released FlowMap~\cite{flowmap} are the dominant approaches to static scene camera parameter estimation, and RoDynRF~\cite{rodynrf} is the state-of-the-art method for estimating camera parameters for dynamic scenes. Since COLMAP~\cite{colmap} and FlowMap~\cite{flowmap} are both fundamentally developed for static scenes, and as discussed in the limitation section of FlowMap~\cite{flowmap} paper, the estimated camera parameters of FlowMap are less accurate than the ones from COLMAP~\cite{colmap}, in the following subsections we use COLMAP~\cite{colmap} and RoDynRF~\cite{rodynrf} as two baselines for camera parameters estimation; the evaluation setups are discussed in Appendix Sec \ref{implement}.
\vspace{-5pt}
\subsection{Evaluation of Estimated Camera Parameters}
\label{cameraevaluation}
\begin{table}
  \small
  \caption{\textbf{Camera Parameter Prediction Errors for NeRF-DS Dataset.} We use the COLMAP camera parameters as ground truth and show the error measures ATE$\downarrow$/RPR trans$\downarrow$/EPR rot$\downarrow$ for each method.}
  \label{result_nerfds_cam}
  \centering

    \begin{threeparttable}
        \begin{tabular}{>{\centering\arraybackslash}p{1.0cm}
                        >{\centering\arraybackslash}p{1.3cm}
                        >{\centering\arraybackslash}p{1.3cm}
                        >{\centering\arraybackslash}p{1.3cm}
                        >{\centering\arraybackslash}p{1.3cm}
                        >{\centering\arraybackslash}p{1.3cm}
                        >{\centering\arraybackslash}p{1.3cm}
                        >{\centering\arraybackslash}p{1.3cm}}
            \toprule
            Method & bell  &  as &  basin  & plate & press & cup & sieve \\
            \midrule
            % COLMAP & -  & - & - & - & - & - & - \\
            RoDynRF & .14/.18/11.5 & .13/.17/10.8 & .13/.15/11.8 & .13/.18/10.3 & .13/.18/10.5 & .12/.16/11.8 & .14/.18/15.7\\
            Ours & .02/.03/2.05 & .03/.04/3.68 & .02/.03/1.94 & .07/.09/6.86 & .03/.03/3.68 & .01/.01/1.42 & .02/.03/2.57\\
            \bottomrule
        \end{tabular}
    % \begin{tablenotes}\small
    %     \item[a] In this table, we assume the camera poses from COLMAP as G.T., and compare other works with it.
    % \end{tablenotes}
    \end{threeparttable}
\end{table}

% \vspace{-10pt}

\begin{figure}[ht]
    \centering
    \includegraphics[width=\linewidth]{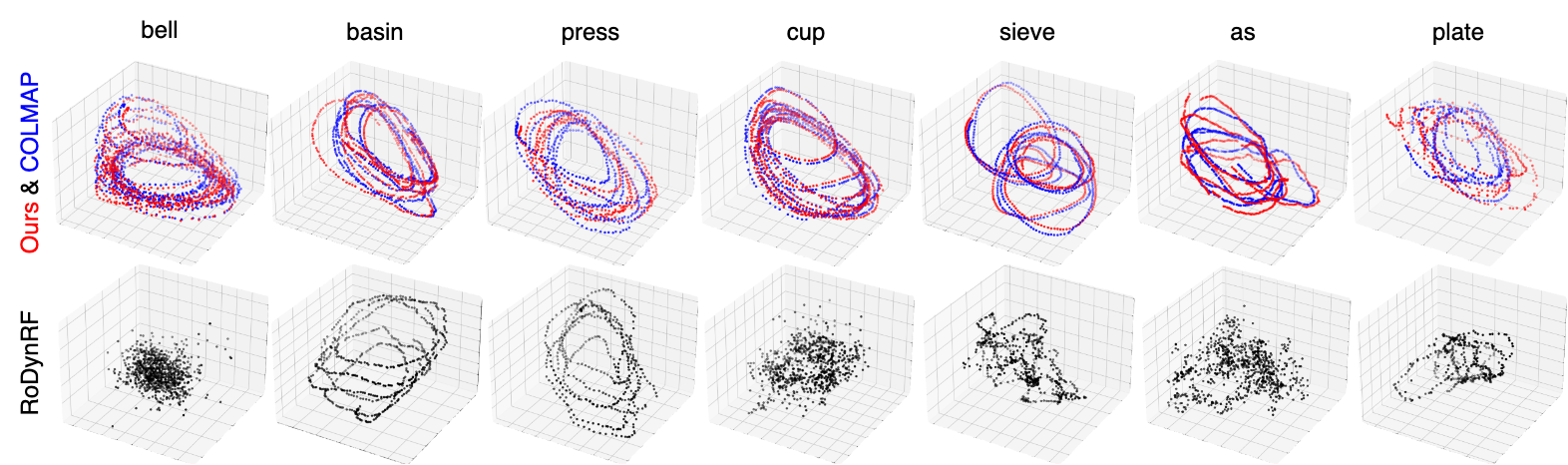}
    \vspace{-10pt}
    \small
    \caption{\textbf{Visual Camera Comparisons on NeRF-DS.} The red \textcolor{red}{\textbullet}, blue \textcolor{blue}{\textbullet}, and black \textcolor{black}{\textbullet} bullets respectively represent the estimated camera poses by our approach, COLMAP~\cite{colmap}, and RodynRF~\cite{rodynrf}.}
    \vspace{-7pt}
    \label{nerfds-colmapvsours}
\end{figure}
\vspace{1pt}
%\textbf{metrics descriptiom-------?????????}

\begin{figure}[ht]
    \centering
    \includegraphics[width=\linewidth]{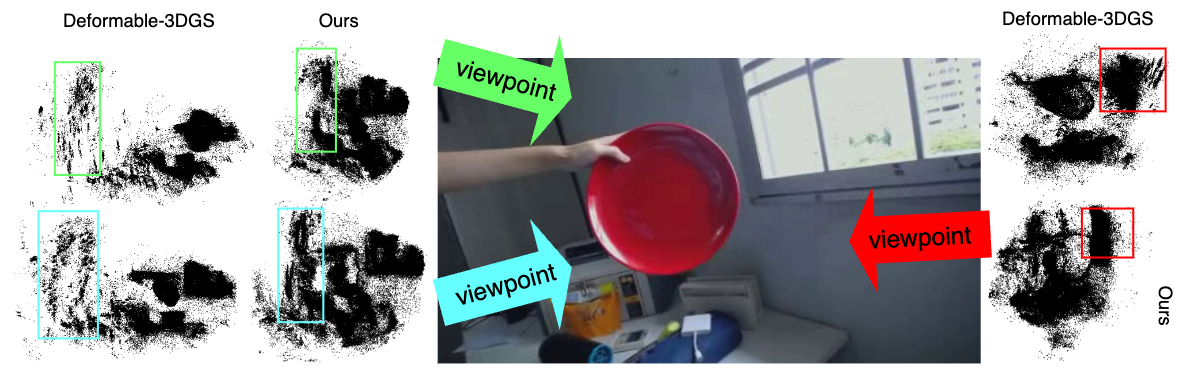}
    \vspace{-20pt}
    \small
    \caption{\textbf{Optimized Point Cloud Comparisons.} We take the \texttt{plate} scene in the NeRF-DS~\cite{nerfds} dataset as the example here and show more in Appendix Sec \ref{morepointcloud}. The boxes and the corresponding viewpoints are color-coded. The dense points due to the back wall plane formed using our estimated camera parameters, shown in the \textcolor{green}{green boxes} and \textcolor{blue}{blue boxes}, can be seen to be more reasonable, in comparison with the scattered points from the same back wall formed using COLMAP camera parameters. Similar comments apply to the \textcolor{red}{red boxes} corresponding to the window points.} 
    % \vspace{-7pt}
    \label{pointcloud-compare}
\end{figure}

We first present quantitative and qualitative evaluations of camera parameter estimates. 
For quantitative evaluation, we use Absolute Trajectory Error (ATE), Relative Pose Error for Translation (RPE Trans), and Relative Pose Error for Rotation (RPE rot), which represent the global discrepancy between two trajectories, errors in translation between consecutive imaging instants (frames), and
errors in orientation between consecutive imaging instants (frames). (See Appendix Sec.~\ref{metric} for details.)
For the NeRF-DS~\cite{nerfds} dataset, containing \textbf{long monocular videos} and \textbf{small object movements} with relatively \textbf{large camera movements}, we show quantitative and visual comparisons in Table \ref{result_nerfds_cam} and Fig \ref{nerfds-colmapvsours}. Our method obtains estimates comparable to COLMAP's ~\cite{colmap}, but RoDynRF~\cite{rodynrf} fails here, showing our method is more accurate than RoDynRF~\cite{rodynrf}. 
% ?????????????I DO NOT UNDERSTAND THE POINTS BEING MADE HERE?????????
Furthermore, we show the optimized point cloud comparisons between Deformable-3DGS~\cite{deformable-3dgs} with COLMAP cameras and our SC-4DGS with our estimated cameras in Fig \ref{pointcloud-compare}.
% Further, the camera pose estimates from COLMAP and ours show an obvious difference in camera trajectory for the \texttt{plate} data. For comparison, we show the optimized point cloud results of Deformable-3DGS~\cite{deformable-3dgs} with the camera from COLMAP and ours with our self-calibration in Fig.\ref{pointcloud-compare}.
% ??????????????
The optimized point clouds of Deformable-3DGS~\cite{deformable-3dgs} contain points that appear to be floating and not a part of scene geometry \texttt{plate}, e.g., the ones in the colored boxes. In contrast, the optimized point clouds from our method are spatially more concentrated and geometrically contiguous. We can use such comparisons to see the misalignments between camera pose estimates from COLMAP and our method and their relative accuracies (Fig \ref{nerfds-colmapvsours}.
For example, we can select the correspondences of a point across three frames and do ray casting from each camera center going through the corresponding frame point. If the camera parameters are accurate, these three rays should intersect at the same 3D point. When the camera parameter estimates have errors, the resulting lack of triangulation relation among these three rays, the RGB loss minimization during rendering will encourage the Adaptive Density Control to add more floating points in each frame. This phenomenon becomes more pronounced as the distance to the camera increases. We present more such comparisons in Appendix Sec \ref{morepointcloud} and Fig \ref{pointcloud-nerfds}. Further, we note that in Tab. \ref{result_nerfds_cam}, the EPR rot errors seem relatively bigger than ATE errors and RPE trans errors. The reason is that while calculating EPR rot, the differences between the scales of our learned camera coordinates and COLMAP camera coordinates amplify the EPR rot errors, without affecting the accuracy of our optimized camera parameters. The ATE and RPE trans metrics in Tab. \ref{result_nerfds_cam}, visual camera results in Fig \ref{nerfds-colmapvsours}, and the rendering results in Fig \ref{render-nerfds} help bring out the effectiveness of our method.

\vspace{-3pt}

We next demonstrate the better performance of our method over COLMAP and RoDynRF~\cite{rodynrf} for the case of dynamic scenes with relatively \textbf{large object movements} and \textbf{tiny camera translations} but \textbf{huge camera rotations}.
% such as DAVIS~\cite{davis}. 
In our experiments on the DAVIS~\cite{davis} dataset, we find that COLMAP~\cite{colmap} fails on more than 80\% of the scenes in DAVIS~\cite{davis} even when we provide the ground truth motion mask. This finding is also consistent with the observations made in RoDynRF~\cite{rodynrf}. In investigating why COLMAP fails, we find that it successfully extracts features in the first step but the failures start from the exhaustive feature matching step.

\begin{figure}[ht]
    \centering
    \includegraphics[width=\linewidth]{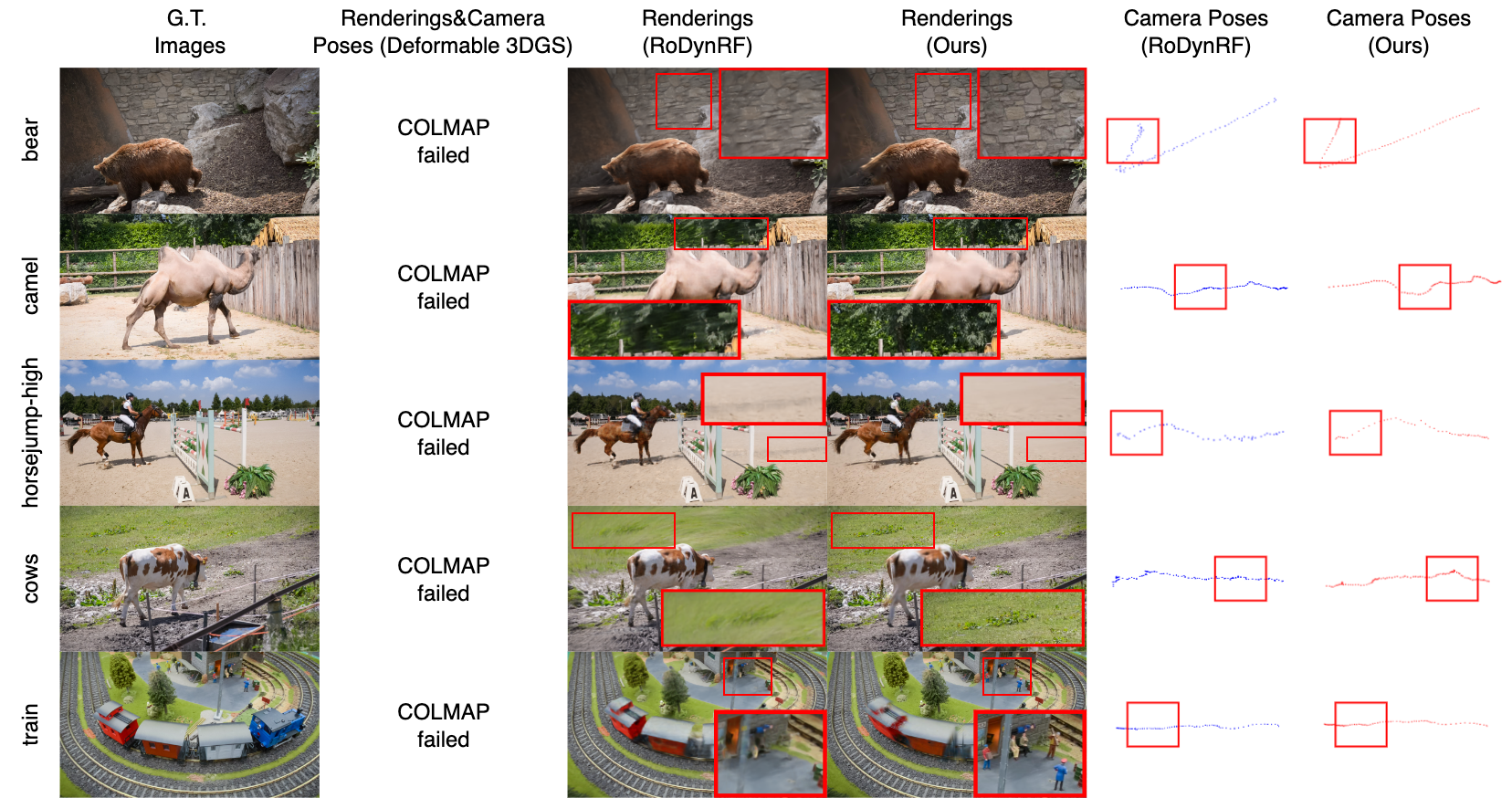}
    \vspace{-10pt}
    \small
    \caption{\textbf{Rendering \& Camera Pose Comparisons on DAVIS.} For each scene, we show the camera pose comparisons and rendering comparisons among Deformable-3DGS~\cite{deformable-3dgs}, RoDynRF~\cite{rodynrf} and ours, marking the relatively large pose or rendering differences with \textcolor{red}{red boxes.}}
    %
    % We mark the areas with the comparatively ??????????????????large differences in camera poses and renderings???????????????????????????????????????????????????????? between ours and RoDynRF~\cite{rodynrf} with \textcolor{red}{red box}.}
    %
    \vspace{0pt}
    \label{davis-all}
\end{figure}

As shown in Fig.~\ref{davis-all}, COLMAP~\cite{colmap} fails on such scenes, so the methods relying on COLMAP like Deformable-3DGS~\cite{deformable-3dgs} also fail. By contrast, RodynRF~\cite{rodynrf} and our method can provide reasonable camera parameter results and good rendering performance. In our experiments, we found that in most cases, the camera poses estimated by our method and RoDynRF are similar and lead to high-quality renderings, demonstrating that both learn good scene representations. However, in some cases, there are obvious differences between the camera poses estimated by our SC-4DGS and RoDynRF. In these cases, we observe that our renderings exhibit higher fidelity and detail (shown in the middle of Fig \ref{davis-all}). This suggests that our method can estimate camera poses more reliably.

%
% As shown in Fig.\ref{davis-all}, we mark the camera poses with significant difference in red boxes. From the marked areas on camera poses, we randomly select the same frames (cameras) and display the rendering results in the middle of Fig.\ref{davis-all}. Since COLMAP~\cite{colmap} completely fails on such scenes, the methods relying on COLMAP like Deformable-3DGS~\cite{deformable-3dgs} also definitely fail. By contrast, in Fig.\ref{davis-all}, RodynRF~\cite{rodynrf} and our method can provide reasonable camera parameter results and good rendering performance. However, despite the renderings being almost identical in most frames, in frames with significant differences in camera poses, our renderings contain clearer details, demonstrating our more accurate camera parameters. In addition, we also show the camera pose comparisons between ours and COLMAP\cite{colmap} on the \textbf{non-video} Nvidia~\cite{nvidia} dataset in Appendix Sec.\ref{cameranvidia}.
% ??????????????????????????????????????????????????????????

\vspace{-5pt}
\subsection{Rendering Evaluation}

\begin{figure}[ht]
    \centering
    \includegraphics[width=\linewidth]{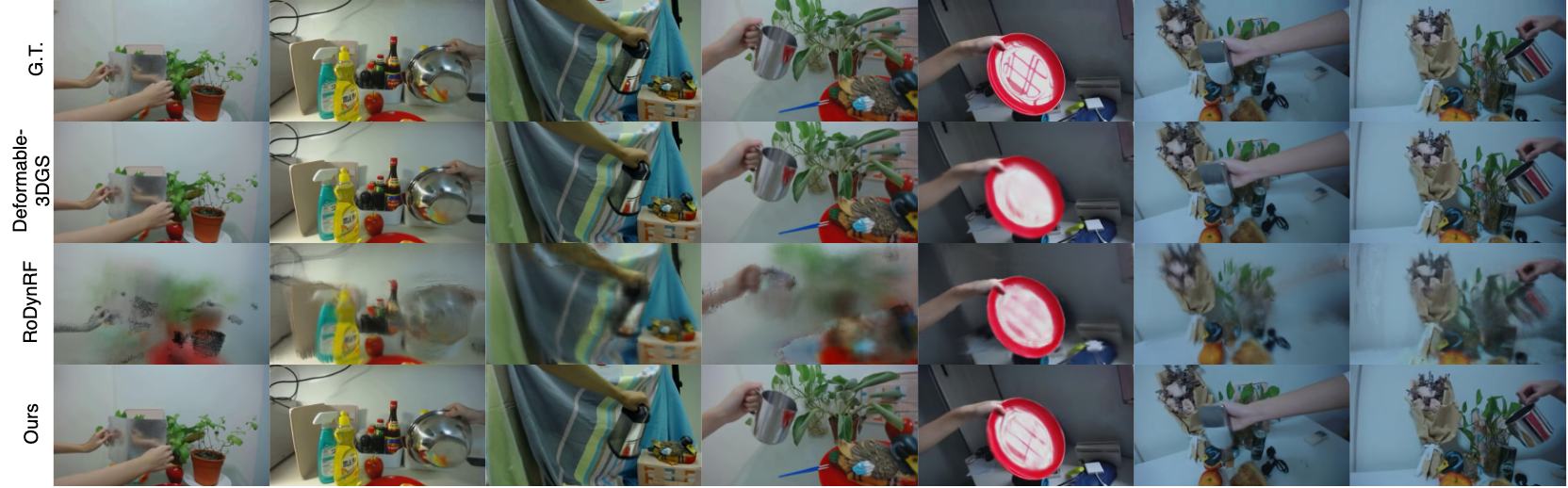}
    \vspace{-10pt}
    \small
    \caption{\textbf{Visual Novel View Synthesis Results on NeRF-DS.}}
    \vspace{-7pt}
    \label{render-nerfds}
\end{figure}

We evaluate the performance of novel view synthesis through quantitative results and qualitative comparisons in Tab \ref{result_nerfds} and Fig \ref{render-nerfds}. Due to the inability to estimate accurate camera parameters from long videos, RoDynRF~\cite{rodynrf} cannot learn good dynamic scene representations. Therefore, it cannot render high-quality images from novel views. Its PSNR and SSIM are significantly lower than the others' by approximately 10.00 and 0.20, representing low image-reconstruct quality. Its LPIPS is higher than the others' by around 0.35, meaning more noise. The rendered novel view frames in Fig \ref{render-nerfds} often have blur, floating points, or other noise. Regarding the rendering comparisons between ours and Deformable-3DGS~\cite{deformable-3dgs}, we use the outpus from COLMAP~\cite{colmap} in all experiments of Deformable-3DGS, and our calibration results in all experiments of our method. Under our self-calibration, our SC-4DGS can achieve comparable PSNR, SSIM, and LPIPS with the Deformable-3DGS~\cite{deformable-3dgs}, and even have high-quality rendering details in some rendering frames. For example, although the PSNR of Deformable-3DGS on the \texttt{plate} scene is 0.005 better than ours, as shown in the fifth column of Fig \ref{render-nerfds}, our method can render more detailed lighting and shadow effects on the plate. The reason behind this is that due to the less inaccurate camera parameters from COLMAP~\cite{colmap}, the Adaptive Density Control adds more points to the scene to adapt to the RGB loss, as shown in Fig \ref{pointcloud-compare}. Such floating points might help improve NVS performance at the viewpoints near the training views; however, because of the wrong geometry, the NVS performance at the viewpoints far from the training views is extremely poor.

\begin{table} 
  \small
  \caption{\textbf{Rendering Results: PSNR $\uparrow$/SSIM$\uparrow$/LPIPS$\downarrow$ for NeRF-DS.}}
  \label{result_nerfds}
  \centering
  \begin{threeparttable}
      \scalebox{0.99}{
      \begin{tabular}{ccccccccc}
        \toprule
        Method  & Metric & bell & as & basin & plate & press & cup & sieve\\
        \midrule
        \multirow{3}{*}{\begin{tabular}[c]{@{}c@{}}Deformable \\ \small{3DGS}~\cite{deformable-3dgs}\end{tabular}} & PSNR  & 31.9745 & 36.7205 & 33.1977 & 30.0201 & 37.1836 & 36.4271 & 37.2177\\
                              & SSIM  & 0.9297 & 0.9629 & 0.9450 & 0.9141 & 0.9642 & 0.9621 & 0.9692\\
                              & LPIPS  & 0.1174 & 0.0927 & 0.0974 & 0.1304 & 0.0980 & 0.0854 & 0.0801\\
        \midrule
        \multirow{3}{*}{RoDynRF~\cite{rodynrf}} & PSNR  & 22.7290 & 20.9097 & 20.3676 & 25.4060 & 22.5965 & 20.9977 & 28.2272\\
                              & SSIM  & 0.7018 & 0.7409 & 0.6845 & 0.8230 & 0.7612 & 0.6791 & 0.8537\\
                              & LPIPS  & 0.3959 & 0.4411 & 0.4201 & 0.2819 & 0.3971 & 0.4891 & 0.2695\\
        \midrule
        \multirow{3}{*}{Ours} & PSNR  & 31.2146 & 36.4721 & 32.9480 & 30.0175 & 35.8331 & 35.9480 & 36.9229\\
                              & SSIM  & 0.9240 & 0.9611 & 0.9416 & 0.9140 & 0.9535 & 0.9590 & 0.9668\\
                              & LPIPS  & 0.1231 & 0.0879 & 0.1001 & 0.1318 & 0.1007 & 0.0841 & 0.0755\\
        \bottomrule         
      \end{tabular}
      }
    \end{threeparttable}
\end{table}

\begin{figure}[ht]
    \centering
    \includegraphics[width=\linewidth]{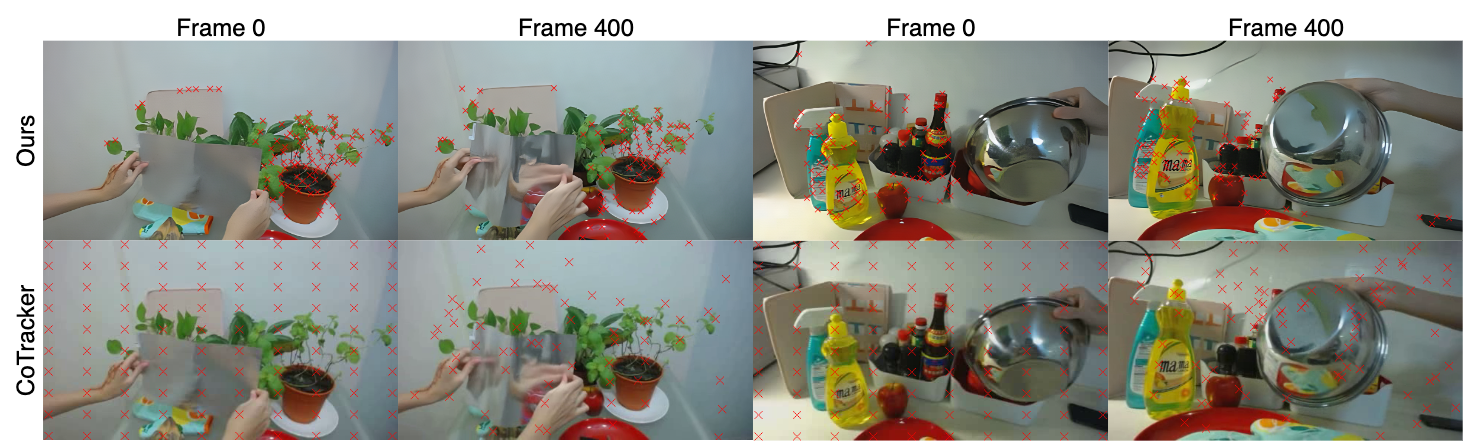}
    \vspace{-10pt}
    \small
    \caption{\textbf{Quality of point tracking by SPE (Ours) vs CoTracker for NeRF-DS.}}
    \vspace{-7pt}
    \label{ctspe}
\end{figure}
\subsection{Why we do not use the CoTracker output directly - Ablation Study}
\label{ablation}

%\textbf{Why do we not use CoTracker output directly?} 
Our SPE algorithm is based on the point tracking (PT) method used in CoTracker~\cite{cotracker}. This tracking helps obtain correspondence between $\mathbf{SP}_{i}^{3D}$ and $\mathbf{SP}_{i}^{2D}$. Since it is difficult for almost all state-of-the-art dense PT models, e.g.,~\cite{cotracker,omnimotion}, to accurately track every point in every frame in each scene, we do not fully trust the results from any off-the-shelf models unlike CF-3DGS~\cite{cf-3dgs}, RoDynRF~\cite{rodynrf} and FlowMap~\cite{flowmap}. In Fig \ref{ctspe}, we show that the direct implementation of CoTracker~\cite{cotracker} results in major point-tracking errors. 
The red \textcolor{red}{$\times$} denotes the location of the selected structural point in frame 0, and of the same structural point after tracking in frame 400. 
%
% The red \textcolor{red}{$\times$} in Frame 0 and Frame 400 respectively denotes the 2D locations of $\mathbf{SP}_{0}^{2D}$ and $\mathbf{SP}_{400}^{2D}$ obtained from our proposed SPE algorithm.
%
Even the state-of-the-art dense point-tracking model CoTracker cannot correctly track points in low-texture background regions like walls, due to their highly similar features. Also in high-texture foreground regions such as leaves, numerous points exhibit confusingly similar features. Alternatively, our results in Fig \ref{ctspe} show that the SPE algorithm can filter out the most reliable point-tracking results as 2D structural points. If some points become invisible during tracking, the automatic structural point adaptation mechanism introduces new structural points for tracking to continue estimation of frame-to-frame relationships. For space reasons, here we show only qualitative results for two dynamic scenes in the NeRF-DS dataset; Appendix Sec \ref{moreab} contains more comparisons.

\vspace{-5pt}
\section{Conclusions}
\vspace{-5pt}
In this paper, we propose \textbf{SC-4DGS} for 4D novel view synthesis w/o camera priors. Our experiments demonstrate that our approach yields more reliable and robust estimates of camera parameters than the state-of-the-art for videos of varying lengths and scenarios, particularly for extreme geometry scenes; obtains optimal dynamic scene representations, and synthesizes high-fidelity RGB images from novel views. We believe that our camera parameter estimation algorithm may also benefit other tasks requiring camera self-calibration. 

\textbf{Limitations.} A limitation that needs to be overcome is our underlying assumption that the focal length remains constant across frames; if overcome, it would allow self-calibrating 4D NVS with variable zoom effects.
Another limitation is that our method requires ground truth motion masks as input which becomes difficult to specify for scenes containing areas of high-speed fluid motion.

\textbf{Acknowledgements.} This work was supported by the Office of Naval Research (grant N00014-20-1-2444) and the USDA National Institute of Food and Agriculture (grant 2020-67021-32799 /1024178). Support for computational resources was provided by the National Science Foundation (awardOAC 2005572) and the State of Illinois, through Delta, a joint effort of the University of Illinois Urbana-Champaign and its National Center for Supercomputing Applications.

% \newpage
\clearpage
\bibliographystyle{plain}
\bibliography{ref}

%%%%%%%%%%%%%%%%%%%%%%%%%%%%%%%%%%%%%%%%%%%%%%%%%%%%%%%%%%%%
\newpage
\appendix
\label{appendix}
\section{Appendix}

\subsection{Datasets}
As the novel view synthesis tasks highly rely on the estimated camera parameters from COLMAP~\cite{colmap}, the NVS datasets will be released only if the contributors can make sure that COLMAP~\cite{colmap} can estimate the relatively accurate camera parameters in each scene. Due to such reasons, besides NeRF-DS~\cite{nerfds} and Nvidia~\cite{nvidia}, following RoDynRF~\cite{rodynrf}, we also evaluate our results on the DAVIS~\cite{davis} dataset which includes more general wild scenarios with large movements and deformation. In summary, we test our methods on challenging datasets with large object movements and large camera movements. The scenarios include indoor scenes, outdoor urban scenes, and outdoor wild scenes.

\textbf{NeRF-DS.} The NeRF-DS~\cite{nerfds} dataset comprises seven monocular long videos from seven distinct dynamic scenes, each containing between 400 and 800 frames. Every scene includes at least one specular object and features a mix of low-texture and high-texture backgrounds. Furthermore, the dataset exhibits significant scene and camera movements. Due to some blurriness in the provided frames, we have applied the RealBasicVSR~\cite{realbasicvsr} model for deblurring. To ensure fairness, deblurred frames are used consistently across all experiments and ablation studies involving the NeRF-DS~\cite{nerfds} dataset. The ground truth RGB images and motion masks are provided. Camera parameters are determined using COLMAP~\cite{colmap}. In line with other studies~\cite{deformable-3dgs, nerfds}, we utilize the highest resolution images available ($480 \times 270$) for scene optimization.

\textbf{DAVIS.} DAVIS~\cite{davis} dataset has 40 short monocular video sequences of 40 different dynamic large scenes including animals, humans, vehicles, etc. Each video sequence contains 50 - 80 frames with at least one moving object with the ground truth RGB images and motion masks. Following RoDynRF~\cite{rodynrf}, we choose the challenging sequences with relatively large camera and objects' movements for comparisons and use the largest resolution ($1920 \times 1080$) for optimization.

\textbf{Nvidia.} Nvidia~\cite{nvidia} dataset contains 9 dynamic scenes. Each scene is recorded by 12 cameras at 12 timestamps and at each timestamp, each camera takes one image. For each scene, the input monocular video is made up of selecting one image taken by one camera at one timestamp without duplications. More detail can be referred to ~\cite{nvidia}. In the Nvidia dataset, the camera parameters are preprocessed by COLMAP~\cite{colmap}, and the foreground masks are given. We use the default resolution ($960 \times 540$) for opimization.

\label{moreab}
\begin{figure}[ht]
    \centering
    \includegraphics[width=\linewidth]{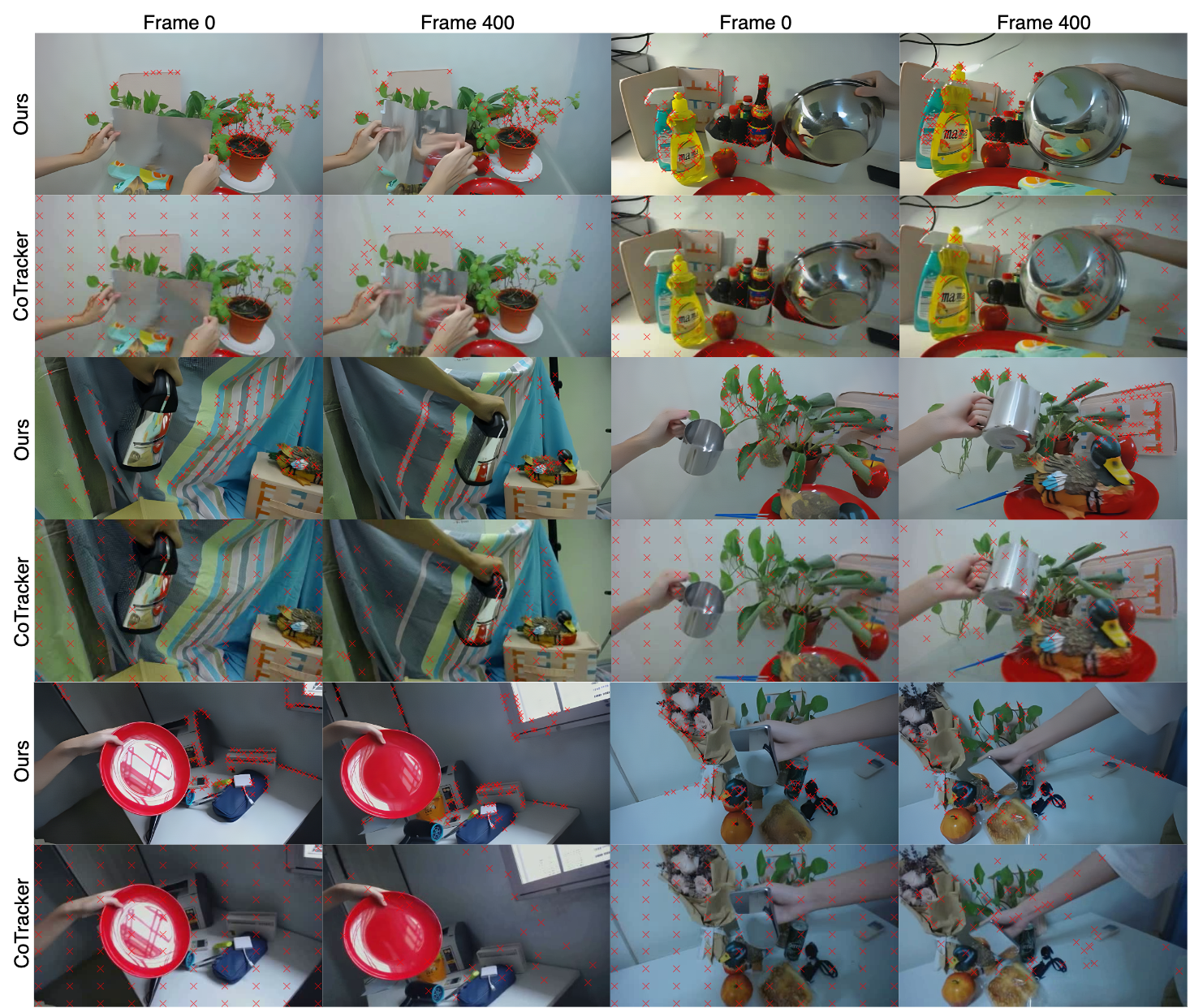}
    \vspace{-10pt}
    \small
    \caption{\textbf{More Ablation Results on SPE v.s. CoTracker.}}
    \vspace{-7pt}
    \label{ctspe2}
\end{figure}

\subsection{Implementation Details}
\label{implement}
\textbf{Device.} All experiments in this paper are conducted on one NVIDIA A100 40GB GPU.

\textbf{Model Setup.} During the camera parameters \& 3D structural points optimization period, we parameterize quaternion, translation, and focal length for each camera, and $H$ learnable 3D structural points. We also adopt the constant learning rates equaling 0.01, 0.01, 1.0, and 0.01 for quaternion, translation, focal length, and the positions of 3D structural points. In the overall detailed dynamic scene optimization step, we follow the same model settings with Deformable-3DGS~\cite{deformable-3dgs}.

\textbf{Evaluation Setup.} For camera pose evaluations, following RoDynRF~\cite{rodynrf}, we take COLMAP~\cite{colmap} as the baseline, although sometimes COLMAP~\cite{colmap} cannot work well in the scenes with large movements. In our evaluations of novel view synthesis, we adapt our assessment approaches to accommodate the diverse configurations of various datasets. For example, the NeRF-DS~\cite{nerfds} dataset provides training videos and testing videos, however, the estimated camera parameters from different works on different videos can be in different coordinates, and the direct alignment will cause large errors of aligned rotations. Under such circumstances, we split the frames of the training videos in each scene into training sets and testing sets. For every two frames, we incorporate the first frame into the training set and the second frame into the testing set. Besides, since COLMAP~\cite{colmap} always fails on DAVIS~\cite{davis}, following the setup in RoDynRF~\cite{rodynrf}, we only compare the scene representations and camera poses between ours and RoDynRF since no testing sets are available. Besides, the tremendous movements of objects across frames make Although our approach is developed for monocular video input, we also show the camera pose comparisons between ours and COLMAP~\cite{colmap} on the non-video Nvidia~\cite{nvidia} dataset.

% \textbf{iPhone.} iPhone\cite{iphone} dataset has 14 long monocular video sequences of 14 challenging dynamic scenes with 

\subsection{Evaluation Metrics}
\label{metric}
We follow the same camera and rendering evaluation metrics as the existing works~\cite{rodynrf}.
\subsubsection{Rendering Evaluation Metrics}
\textbf{PSNR.} PSNR is a widely used metric for measuring the quality of reconstructed images compared to ground truth images. It is expressed in decibels (dB). The higher the PSNR, the better the quality of the reconstructed image. PSNR is calculated using the mean squared error (MSE) between the original and the reconstructed image:

\begin{equation}
\label{psnr}
\vspace{-0pt}
\begin{split}
PNSR = 10 \cdot \log_{10} \left(\frac{MAX^2}{MSE(Image_{Recon}, Image_{G.T.})}\right),
\end{split}
\vspace{-0pt}
\end{equation}

where $MAX$ is the maximum possible pixel value of the image.

\textbf{SSIM.} SSIM is designed for assessing the perceived quality of digital images and videos. SSIM focuses on the changes in structural information, luminance, and contrast. The values of SSIM range from -1 to 1, where 1 represents perfect. The formula is defined as:
\begin{equation}
\label{ssim}
\vspace{-0pt}
\begin{split}
SSIM = \frac{(2\mu_x \mu_y + c_1)(2\sigma_{xy} + c_2)}{(\mu_x^2 + \mu_y^2 + c_1)(\sigma_x^2 + \sigma_y^2 + c_2)},
\end{split}
\vspace{-0pt}
\end{equation}

where $x$ and $y$ represent two images, $\mu_{x}$ and $\mu_{y}$ are the average of $x$ and $y$. $\sigma_{x}^{2}$ and $\sigma_{y}^{2}$ are variance of $x$ and $y$. $\sigma_{xy}$ stands for the covariance of $x$ and $y$, and $c_1$, $c_2$ are the regularization factors.

\textbf{LPIPS.} LPIPS is a recent metric measuring the perceptual difference between two images as perceived by a trained neural network. In this paper, we use the same trained network with existing works~\cite{3dgs, rodynrf, deformable-3dgs}. LPIPS quantifies the difference in the feature representations of images within the neural network, suggesting that a higher LPIPS score indicates a greater perceptual difference.

\subsubsection{Camera Evaluation Metrics}
\textbf{Absolute Trajectory Error (ATE).} ATE measures the discrepancy between the true trajectory and the estimated trajectory that a robot or a camera follows over a period of time. It provides a global error measurement over the entire trajectory. ATE is computed by aligning the estimated trajectory with the ground truth trajectory and then computing the Euclidean distances between corresponding points on the aligned trajectories.

\textbf{Relative Pose Error for Translation (RPE Trans).} RPE Trans measures the error in the translation part of the pose between consecutive poses or over a fixed time/distance interval. Unlike ATE, RPE focuses on the local accuracy of the motion estimation, examining how well the system preserves the relative motion between two points in time or space.

\textbf{Relative Pose Error for Rotation (RPE Rot).} RPE Rot measures the error in the orientation between estimated poses relative to the true orientation. This metric is computed by determining the difference in orientation between the estimated and ground truth poses over short sequences and is typically expressed in angular units (like degrees or radians).

% \subsection{More Experiment Results}
% \subsubsection{Qualitative Results on \textcolor{red}{NeRF-DS}\cite{nerfds} dataset}
% aaa
% \subsubsection{Qualitative Results on \textcolor{red}{Nvidia}\cite{nvidia} dataset}
% aa
% \subsubsection{Qualitative Results on \textcolor{red}{DAVIS}\cite{davis} dataset}
% aa

\subsection{More Ablation Study Results}
\textbf{SPE v.s. CoTracker.} Here in Fig \ref{ctspe2} we show more comparisons between the results from the direct implementation of CoTracker~\cite{cotracker} and the results from our introduced SPE algorithm. The comparisons are conducted on the NeRF-DS~\cite{nerfds} dataset. As there are two scenes with completely the same background, we only show one of them here.

\subsection{Camera Comparisons on Nvidia dataset}
\label{cameranvidia}
\begin{figure}[ht]
    \centering
    \includegraphics[width=\linewidth]{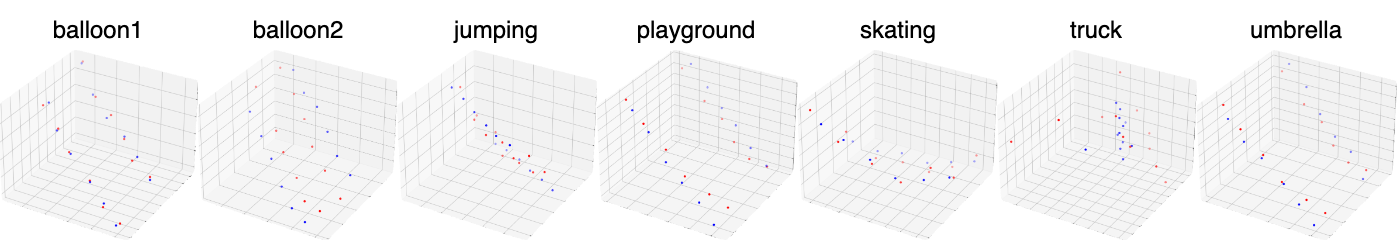}
    \vspace{-10pt}
    \small
    \caption{\textbf{Visual Camera Comparisons on Nvidia.} The red \textcolor{red}{\textbullet} and blue \textcolor{blue}{\textbullet} respectively represent the estimated camera poses from our SC-4DGS and COLMAP~\cite{colmap}}
    \vspace{-7pt}
    \label{nvidia-camera}
\end{figure}

Our SC-4DGS model requires the monocular videos as input, and one set of individual images as input might somehow degrade the performance of our method. Despite this, we still test our method and COLMAP~\cite{colmap} on the non-video Nvidia~\cite{nvidia} dataset, as shown in Fig \ref{nvidia-camera}. In most scenes, our method can still produce comparable results with COLMAP~\cite{colmap}.

\subsection{More Optimized Scene Point Clouds Comparisons}
\label{morepointcloud}
\begin{figure}[ht]
    \centering
    \includegraphics[width=\linewidth]{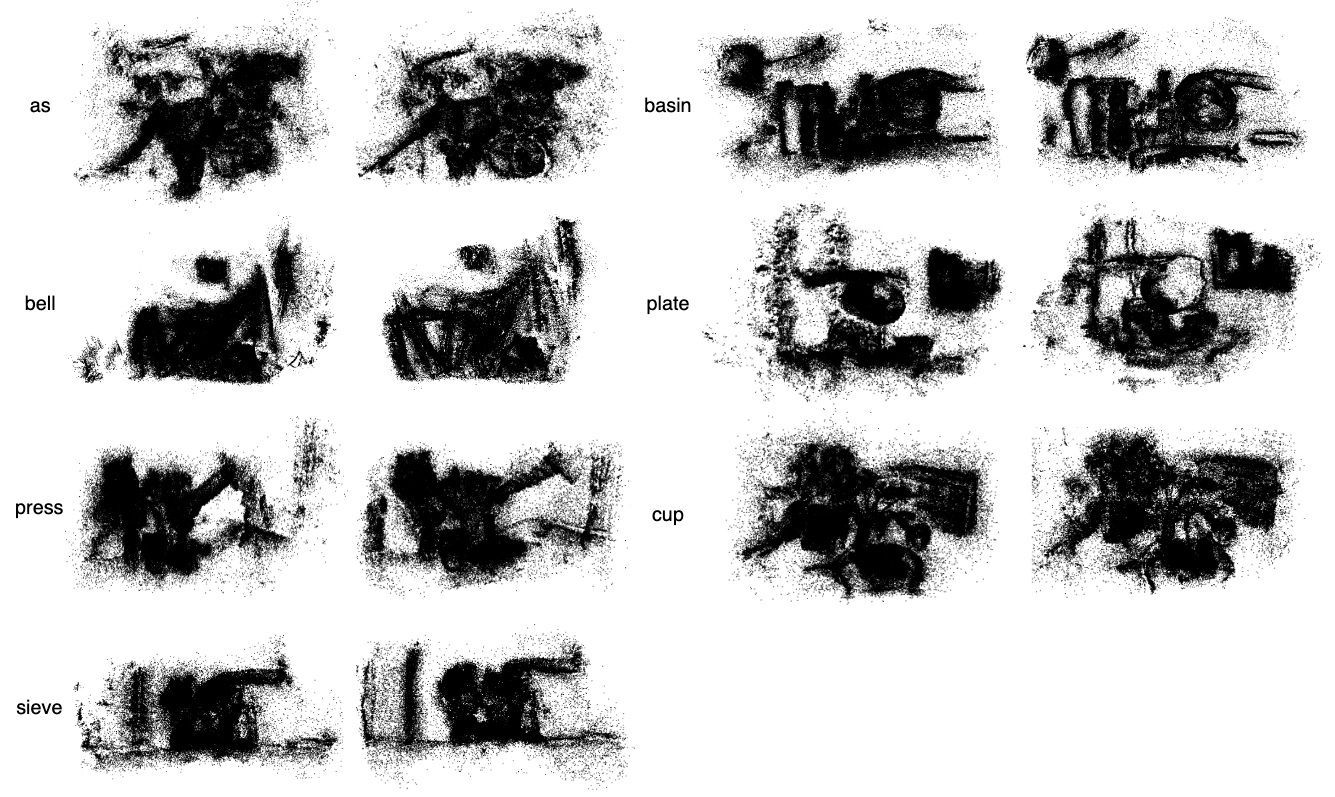}
    \vspace{-10pt}
    \small
    \caption{\textbf{Optimized Scene Point Clouds on NeRF-DS.} For each scene, the left figure is from Deformable 3DGS~\cite{deformable-3dgs}, the right figure is from Our method.}
    \vspace{-7pt}
    \label{pointcloud-nerfds}
\end{figure}

In Fig \ref{pointcloud-nerfds}, we show more optimized point cloud results of the scenes in the NeRF-DS~\cite{nerfds} dataset. Among these comparisons, Deformable-3DGS~\cite{deformable-3dgs} prefers to add more floating points to adapt to the RGB loss. However, these floating points cannot represent the real geometry of the scenes. This can promote the accuracy of the novel view synthesis from easy test viewpoints like the ones around the training viewpoints. However, it will destroy the real geometry of scenes, leading to poor NVS performance at the views that are not around the training viewpoints.

\subsection{Structural Point Extraction}
We show the detailed SPE algorithm in Alg \ref{a1}.

\begin{algorithm}
\caption{Structural Points Extraction (SPE)}
\begin{algorithmic}[1]
\State Initialize $H = 0, \mathbf{P}^{pos} \in \mathds{-1}^{N\times\tau\times2}, \mathbf{P}^{index}\in \mathds{-1}^{N\times\tau}$.
\For{$i = 0, 1, 2, \dots, N - 1$}
    \If{$-1 \texttt{ in } \mathbf{P}_{i}^{index}$}:
        \State $\mathbf{E}_{i}^{gray} \leftarrow \text{Canny Edge Detector}(\mathbf{F}_{i}^{gray})$
        \State $\mathbf{Grad}_{i}^{magni} \leftarrow Sqrt(Sum(\mathbf{Grad}_{i}^{r}, \mathbf{Grad}_{i}^{g}, \mathbf{Grad}_{i}^{b}))$
        \State  $\mathbf{SP}_{i}^{Pool} \leftarrow \text{Maximum Filter}(\mathbf{E}_{i}^{gray} \cap \mathbf{Grad}_{i}^{magni}, \mathcal{W}) \cap \mathbf{M}_{i}^{motion}$
        \State $credit = \mathds{1}^{B}, p = i + 1$
        \State $\mathbf{Pred}^{pos}, \mathbf{Pred}^{vis} = \text{CoTracker}(\mathbf{F}_{j, (j > i)}^{rgb}, \mathbf{SP}_{i}^{Pool})$
        \While{$p \leq N - 1$}
            \State $credit[\mathbf{Pred}_{p}^{vis} \texttt{==} 0 \cup \mathbf{M}_{p}^{motion}[\mathbf{Pred}_{p}^{pos}] \texttt{==} 0] = 0$
            \If{$\sum(credit_{p - 1}) > num$ \& $\sum(credit_{p}) < num$}
                \State $\mathbf{P}_{m}^{index}, m \in [i, p - 1] \leftarrow [H, H+num)$
                \Comment{Assign index}
                \State $\mathbf{P}_{m}^{pos}, m \in [i, p - 1] \leftarrow \mathbf{Pred}_{m}^{pos}[Random(where(credit_{p - 1} \texttt{==} 1), num)]$
            \EndIf
            \If{$\sum(credit_{p - 1}) < num$ \& $\sum(credit_{p}) < num$}
                \State $\mathbf{P}_{p}^{index} \leftarrow [H, H+num)[credit_{p}[Random(where(credit_{p - 1} \texttt{==} 1), num)] \texttt{==} 1]$
                \State $\mathbf{P}_{p}^{pos} \leftarrow \mathbf{Pred}_{p}^{pos}[where(credit_{p} \texttt{==} 1)]$
            \EndIf
            \State $p \texttt{ += } 1, H\texttt{ += }num$
        \EndWhile
    \Else \texttt{ PASS }
    \EndIf
\EndFor
\State $\mathbf{SP}_{i}^{2D} \leftarrow \mathbf{P}_{i}^{pos}, \mathbf{SP}_{i}^{Index} \leftarrow \mathbf{P}_{i}^{index}(i \in [0, N - 1]), \mathbf{SP}^{3D} \leftarrow \frac{1}{2}\mathds{1}^{H \times 3}$

\end{algorithmic}
\label{a1}
\end{algorithm}

\end{document}